\documentclass[conference]{IEEEtran}
\IEEEoverridecommandlockouts
\IEEEpubid{\makebox[\columnwidth]{https://doi.org/10.1109/COMSNETS59351.2024.10427470 \copyright2024 IEEE \hfill}
\hspace{\columnsep}\makebox[\columnwidth]{ }}
\usepackage{cite}
\usepackage{amsmath,amssymb,amsfonts}
\usepackage{algorithm,algorithmicx,algpseudocode}
\usepackage{graphicx}
\usepackage{textcomp}
\usepackage{xcolor}
\usepackage{dashbox}
\def\BibTeX{{\rm B\kern-.05em{\sc i\kern-.025em b}\kern-.08em
    T\kern-.1667em\lower.7ex\hbox{E}\kern-.125emX}}
\begin{document}

\title{pFedGame - Decentralized Federated Learning using Game Theory in Dynamic Topology\\
}

\author{\IEEEauthorblockN{Monik Raj Behera}
\IEEEauthorblockA{\textit{Department of Computer Science and Engineering} \\
\textit{Indian Institute of Technology, Jodhpur}\\
India \\
behera.3@iitj.ac.in}
\and
\IEEEauthorblockN{Suchetana Chakraborty}
\IEEEauthorblockA{\textit{Department of Computer Science and Engineering} \\
\textit{Indian Institute of Technology, Jodhpur}\\
India \\
suchetana@iitj.ac.in}
}

\maketitle

\begin{abstract}
Conventional federated learning frameworks suffer from several challenges including performance bottlenecks at the central aggregation server, data bias, poor model convergence, and exposure to model poisoning attacks, and limited trust in the centralized infrastructure.
In the current paper, a novel game theory-based approach called `pFedGame' is proposed for decentralized federated learning, best suitable for temporally dynamic networks.
The proposed algorithm works without any centralized server for aggregation and incorporates the problem of vanishing gradients and poor convergence over temporally dynamic topology among federated learning participants.
The solution comprises two sequential steps in every federated learning round, for every participant. 
First, it selects suitable peers for collaboration in federated learning.
Secondly, it executes a two-player constant sum cooperative game to reach convergence by applying an optimal federated learning aggregation strategy.
Experiments performed to assess the performance of pFedGame in comparison to existing methods in decentralized federated learning have shown promising results with accuracy higher than $70\%$ for heterogeneous data.
\end{abstract}


\section{Introduction}
Federated learning (\textbf{FL}) \cite{konevcny2016federated} is a distributed and privacy-preserving machine learning (ML) paradigm, which facilitates collaborative training of a model among participants (\textit{clients or agents}) without any requirement of sharing the data with each other.
FL comprises a `central aggregation server' and `participants' (clients who participate in the federated learning with other participants).
Participants are the actual data owners.
The centralized aggregation server is responsible for the aggregation of participants' models into a global model, which can potentially work for a given participant on a data distribution trained by some other participant.
Zero data exposure from the participant's system to other participants or aggregation server makes FL one of the most practical \textit{privacy preserving machine learning} methodologies in research and industry, such as finance, health care, insurance, Internet of Things, and others.

Centralized aggregation typically suffers from poor convergence on extremely heterogeneous data\cite{tan2022towards}, overgeneralization, model poison attacks and backdoor attacks\cite{fang2020local}, inference attacks\cite{geiping2020inverting}, trust and dependency on central aggregation server and dynamic nature\cite{lim2021decentralized} of the federated learning network.
For the majority of the FL algorithms, the assumption is to have a static network topology, with a consistent communication medium between the participants and aggregation server. 
The above challenges, which impact the real-world adaption of FL are the basis and motivation for the current work.

As discussed in the literature\cite{lim2021decentralized, ye2023personalized}, decentralized FL models are generally resilient to model poisoning attacks, in the absence of a centralized server for potential moderation.
Typically, they follow two-step aggregation in each FL round - a global aggregation by the central server, followed by a localized optimization or fine-tuning, based on stochastic methods.
The related work mainly focuses on global learning with fine-tuning at the local level, ensuring all the participants contribute to the federated learning model, however minuscule the contribution is.
Whereas, the solution discussed in the current work performs a localized FL aggregation, one time, in every FL round for each participant.
This makes the current work localized from the very beginning.

Our contributions to the current work are:
\begin{enumerate}
    \item A two-step decentralized FL approach - `peer selection' and `pFedGame' aggregation, which can work in dynamic network settings without any centralized aggregating server.
    \item A novel, game theoretic-based FL aggregation algorithm, `\textbf{pFedGame}', which formulates a two-player coalition game to reach equilibrium in FL aggregation.
    \item For a given participant, decentralized FL also mitigates the risk of overgeneralization and poor convergence on extremely heterogeneous data, along with support for dynamic topology, which is common in real-world scenarios.
\end{enumerate}

\section{Related Works}
Decentralized FL is being actively researched\cite{ye2023personalized, lim2021decentralized}, in order to overcome challenges of poor aggregation, and diminishing gradient issues observed in highly heterogeneous data distribution among participants. 
They are dependent on a model weight optimization step at the local level (personalization), which fine-tunes the global aggregated model to local, participant's data distribution. 
The work has shown promising results in personalized FL space.
The current work builds on the benefits of personalized FL for decentralization in conventional FL algorithms, which addresses the problem of poor convergence and vanishing gradient\cite{taik2020electrical} problem without a central aggregation server.

Game theory has been studied for incentive mechanisms for participants, optimizations and aggregations of FL algorithms.
A game theoretic approach to reach an optimal and stable FL arrangement is discussed in the work~\cite{donahue2021optimality}, where participants form coalitions or clusters to aggregate models. The contribution of participants in FL model aggregation is directly proportional to the respective model's accuracy on local test data. 
This approach motivates our proposed solution to use coalition game theory for such aggregation algorithms, which can adapt to dynamic networks, without any prior stochastic assumptions of participants in the network.

\section{Problem Statement}

\subsection{System Model and Assumptions}

Federated learning, as a collaborative and distributed machine learning system, can be modeled as a graph $\mathcal{G}(V,E)$, where the \textit{nodes} (denoted by $V$) represent participants and \textit{edges} (denoted by $E$) represent the relationship between the connecting nodes, such as spatial distance, cosine similarity between nodes, etc.
Actual peers for collaboration in every FL round for a given node are decided based on `peer selection' from a set of immediate neighbors. 

\begin{figure}
    \centering
    \includegraphics[width=1\columnwidth]{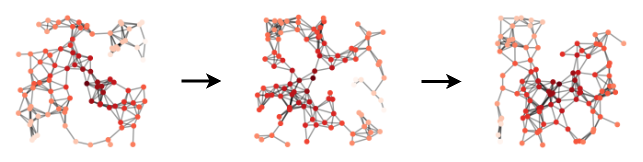}
    \caption{Logical, temporally dynamic representation of graph $\mathcal{G}(V,E)$ over $3$ time steps. Edges represent the relation for every node with peer nodes, based on a certain set of features and properties.}
    \label{fig:dynamic-graph}
\end{figure}

%
Applications of the Internet of Things, Connected Cars and Autonomous Vehicles, or UAV systems generally trace the dynamic network, where nodes are mobile and the relationships among nodes also change with time, resulting in a dynamic graph, as depicted in Fig.~\ref{fig:dynamic-graph}.
As a decentralized system, every node in $V$ is capable of performing federated learning. 
The objective of the paper is to find an optimal FL aggregated model for each node $x \in V$, which collaborates with significant peers in $\mathcal{G}(V,E)$ in the absence of a centralized aggregation server.

Assumptions for the proposed solution are listed below:
\begin{enumerate}
    \item The domain of FL is \textit{horizontal}, where the model across participants follows the same architecture, and the data across participants follows the same schema and feature space.
    \item Every node of $V$ has a bi-directional communication channel with each other, which allows a decentralized mode of communication for FL.
    \item $e \in E$, $e$ is function of \textit{closeness of data distribution} between the corresponding node.
    \item The set of $V$ and $E$ varies over time, which makes $\mathcal{G}(V,E)$ dynamic, topologically and temporally. 
    \item Every node in $V$ can participate in the FL round
    \item For every node, each FL round is supposed to incorporate new learning in an additive manner - the learned weights from the previous FL round combined with learned weights from new data in the current FL round.
\end{enumerate}

\subsection{Mathematical Formulation}

\begin{table}[t]
    \centering
    \begin{tabular}{|l|p{4.5cm}|}
        \hline
        \textbf{Notation} & \textbf{Description} \\
        \hline
        $\mathcal{G}(V,E)$ & Graph with set of vertices and $V$ and set of edges as $E$ \\
        \hline
        $e(a,b)$ & un-directed edge between $a$ and $b$ \\
        \hline
        $C(x)$ & Peer cluster for node $x$ \\
        \hline
        $D(x)$ & Data for node $x$ \\
        \hline
        $M(x)$ & Model for node $x$ \\
        \hline
        $H(m,d)$ & Accuracy of model $m$ on data $d$, range $\in [0, 1]$ \\
        \hline
        $\theta$ & Accuracy threshold for peer selection, range $\in [0, 1]$ \\
        \hline
        $\psi(x)$ & weight of node $x$ in federated learning aggregation \\
        \hline
        $\psi(\alpha)$ & weight of the FL aggregated model from models in $C(x)$ \\
        \hline
        $r$ & number of game rounds for each \textit{pFedGame} aggregation, similar to \textit{epochs in machine learning} \\
        \hline
        $\delta$ & change in $\psi(.)$ for every game rounds, similar to \textit{learning rate in machine learning} \\
        \hline
        $\Gamma(x)$ & FL aggregated model for $x$, using `pFedGame' \\
        \hline
    \end{tabular}
    \vspace{1mm}
    \caption{Mathematical notations}
    \label{tab:math-notation}
\end{table}

Mathematical notations to describe the problem statement are mentioned in Table \ref{tab:math-notation}.
The current solution is a sequential process of `peer selection' and `pFedGame'.
For peer selection, Equation \ref{eq:peer-selection} represents a mathematical condition, which is required for every FL round.

\begin{equation}
    \label{eq:peer-selection}
    C(x) \longleftarrow \bigl\{ c \; | \; H(M(c), D(x)) \geq \theta \; \text{and} \; e(c,x) \in E \bigr\} 
\end{equation}

Global weight update in FedAvg\cite{konevcny2016federated} is discussed in equation \ref{eq:fedavg}.
$w_{t}$ and $w^{k}_{t}$ is the weight of the global model and $k^{th}$ peer's model, respectively, in a FL round $t$. 
$n_{k}$ is the total size of training data in FL round $t$ for peer $k$. $n$ is the sum of the size of training data across all the peers in $V$. 

\begin{equation}
    \label{eq:fedavg}
    w_{t} \longleftarrow \sum_{k=1}^{|V|}\;\dbox{$\frac{n_{k}}{n}$}\;w^{k}_{t}
\end{equation}

\begin{equation}
    \label{eq:fl-func}
    \mathcal{F}([\ldots, M(k)], [\ldots, \psi(k)])
\end{equation}

In the case of `pFedGame', our objective is to replace $\dbox{$\frac{n_{k}}{n}$}$ with $\psi(.)$.
Function to compute aggregation of all the $k$ models using FedAvg\cite{konevcny2016federated} algorithm, with $\psi(.)$ as the weights of peers, used for FL aggregation in FedAvg is described in Equation \ref{eq:fl-func}.
The motivation to do so is inspired by earlier work on personalized FL\cite{fang2020local, geiping2020inverting, lim2021decentralized,ye2023personalized}, which doesn't solely depend on just size of the training data in \textit{FedAvg} algorithm, but considers other statistical parameters, which can reflect potential contribution of aggregated FL model for arbitrary node $x$.
Equation \ref{eq:utility-aggregation} shall converge after finding $\psi(x)$ and $\psi(\alpha)$, which would be ``saddlepoint" in a two-person constant-sum co-operative game.
The cooperative game will be played between two players, which, in a given FL round are (1) the model of node \textit{x}, for which FL aggregation is being done: $M(x)$ and (2) the model resulting from aggregation of models from elements of $C(x)$: $M(\alpha)$.

\begin{equation}
    \label{eq:utility-aggregation}
    max \; H( \; \mathcal{F}([M(x), M(\alpha)], [\psi(x), \psi(\alpha)]), \; D(x)) 
\end{equation}

\section{Methodology}

\subsection{Peer Selection}

Peer selection is the first step in the current work, which is supposed to be executed for every FL round, for every node $\in V$. 
Peer selection restricts the number of nodes (peers), with whom a given node $x$ needs to collaborate during an arbitrary FL round. 
This step essentially decreases the computation that needs to happen during the FL round, by only selecting those peers who can potentially contribute significantly to the given node $x's$ aggregated FL model. 
Algorithm \ref{alg:peer-selection} describes the peer selection methodology proposed in the current work, for a given arbitrary node $x$, which is inspired by the PENS algorithm\cite{onoszko2021decentralized}.
Algorithm \ref{alg:peer-selection} solves for the mathematical objective represented in Equation \ref{eq:peer-selection}.
At the same time, since the algorithm \ref{alg:peer-selection} executes in every FL round for all the nodes, it doesn't assume any pre-defined $\mathcal{G}$ structure, hence, accommodating the dynamic nature of $\mathcal{G}$ over time.

\begin{algorithm}
    \caption{Peer Selection}
    \label{alg:peer-selection}
    \begin{algorithmic}
        \State For an arbitrary node $x$
        \Require $x \in V$, $P$, $C(x) = \phi$, $\theta$
        \Ensure $P \subseteq V$, $\exists p \in P \colon e(x, p) \in E$
        \ForAll{$p \in \left\{ P \cup \{x\} \right\} $}
            \Comment{self-model is also considered for node x}
            \If{$H(M(p), D(x)) \geq \theta$}
                \State $C(x) \leftarrow C(x) \cup \{p\}$
            \EndIf
        \EndFor 
    \end{algorithmic}
    \Return  $C(x)$
\end{algorithm}

\noindent \textit{Complexity Analysis:} As observed from Algorithm \ref{alg:peer-selection}, the time complexity is $O(|P|)$, where the `for' loop runs once over the set $P$. 

Space complexity is $O(|C(x)|)$, which is the resultant set returned from Algorithm \ref{alg:peer-selection}.

\subsection{pFedGame: Game theoretic federated learning aggregation}

`pFedGame' is a novel approach, discussed in the current work, where the intent is to find an optimal $M(x)$ and $M(\alpha)$ for given node $x$, from Equation \ref{eq:utility-aggregation}, using game theory.
`pFedGame' is assumed to be a constant sum, cooperative game between two players, as discussed in \textit{mathematical formulation} subsection earlier.
Every FL aggregation in the FL round, for a given node $x$ is a `game', with few rounds.
Game rounds can be considered similar to `epochs' in machine learning, where the objective of game rounds is to reach a saddle point.

Since the assumption is a constant sum game, Equation \ref{eq:constant-sum} shows the relation between utilities $\psi(x)$ and $\psi(\alpha)$, for two players $M(x)$ and $M(\alpha)$ respectively. 
Algorithm \ref{alg:pFedGame} describes the proposed game theoretic aggregation algorithm for decentralized federated learning.
As a heuristic assumption, initially, $\psi(x) = 0$ and $M(x)$ will perform better on $D(x)$.
Similarly, $\psi(\alpha) = 1$ with $M(\alpha)$ performing worse on $D(x)$.
Higher $\psi(.)$ denotes a higher contribution to FL aggregation.
Assignment of $\psi(x)$ and $\psi(\alpha)$ is contradictory to their initial significance on FL aggregation.
Contradictory initial assignment of $\psi(.)$ is required to facilitate the cooperative game, as described in Algorithm \ref{alg:pFedGame}.

\begin{equation}
    \label{eq:constant-sum}
    \psi(x) + \psi(\alpha) = 1
\end{equation}

\begin{algorithm}
    \caption{pFedGame}
    \label{alg:pFedGame}
    \begin{algorithmic}
        \State For an arbitrary node $x$ in FL round $t$
        \Require $x \in V$, $C(x)$, $\psi(x)$, $\psi(\alpha)$, $r$
        \Comment{$C(x)$ from peer selection}
        \Ensure $|C(x)| \geq 1$, $\psi(x)=0$, $\psi(\alpha)=1$, $r \geq 1$
        \\
        \State $w_{t} \longleftarrow \sum_{k=1}^{|C(x)|}\;\frac{1}{|C(x)|}\;w^{k}_{t}$
        \\
        \State $M(\alpha) \leftarrow w_{t}$
        \State $\beta \leftarrow$  accuracy difference threshold between $M(\alpha)$ and $M(x)$ $\colon$ $\beta$ is empirically significant \\
        \State $\Gamma(x) = \mathcal{F}([M(x), M(\alpha)], [\psi(x), \psi(\alpha)])$
        \For{$i = 1 \rightarrow r$}
            \Comment{$\delta*r \leq 1$}
            \State $\Gamma'(x) = \mathcal{F}([M(x), M(\alpha)], [\psi(x) + \delta, \psi(\alpha) - \delta])$
            \If{$ | H(\Gamma(x), D(x)) - H(\Gamma'(x), D(x)) | \geq \beta $}
                \If{$H(\Gamma(x), D(x)) \leq H(\Gamma'(x), D(x))$}
                \State $\psi(x) \leftarrow \psi(x) + \delta$
                \State $\psi(\alpha) \leftarrow \psi(\alpha) - \delta$
                \State $\Gamma(x) \leftarrow \Gamma'(x)$
                \EndIf
            \EndIf
        \EndFor 
    \end{algorithmic}
    \Return  $\Gamma(x)$
\end{algorithm}

\noindent \textit{Complexity Analysis:} From Algorithm \ref{alg:pFedGame}, it is observed that initial FedAvg aggregation takes $O(|C(x)|)$ time, followed by co-operative game rounds, which takes $O(r)$ time. 
Combined, the time complexity of `pFedGame' is $O(|C(x)|) + O(r)$.

Space complexity of Algorithm \ref{alg:pFedGame} is $O(1)$, since $\Gamma(x), \Gamma'(x), \psi(x), \psi(\alpha)$ variables are being used.

\section{Experiments}

The experiments to validate the proposal of `pFedGame' are conducted in a system with 16GB RAM, 8 physical cores, and 16 virtual threads, Intel $10^{th}$ Gen Core i7 processor with Nvidia GTX 1650Ti 4GB GDDR6 GPU. 
Tensorflow version 2.2 and Numpy were used for neural network and linear algebra framework, which was configured to use parallel threads available to the system within Python version \textit{3.10.6}.
Source code used for the experiments are publicly available\footnote[1]{\textit{https://github.com/bmonikraj/mtp-game-theory-fl/blob/main/MTP\_Core\_LocalFedGT.ipynb}}.

As a baseline comparison to the latest and closest evaluated work\cite{ye2023personalized}, the experiments were performed on 3 different data sets:
\begin{enumerate}
    \item Fashion-MNIST (60,000 grayscale images of 28x28 pixels) with 10 classes\cite{xiao2017fashion}
    \item CIFAR-10 (50,000 images of 32x32 pixels and 3 color channels) with 10 classes\cite{Krizhevsky09learningmultiple}
    \item CIFAR-100 (60,000 images of 32x32 pixels and 3 color channels) with 100 classes\cite{Krizhevsky09learningmultiple}
\end{enumerate}

To evaluate the results of the above data, two separate neural networks were used.
A multi-layer neural network\cite{ye2023personalized} is used for Fashion-MNIST data, whereas a convolution neural network\cite{ye2023personalized} is used for CIFAR-10 and CIFAR-100 data, as described in the `pFedGraph' work.
Similar neural network models have been used in the current experiments to benchmark the performance of `pFedGame' with `pFedGraph'~\cite{ye2023personalized}.
Similar to the simulated federated learning participants settings described in the work\cite{ye2023personalized}, the current experiments have 4 kinds of heterogeneity among simulated participants. 
`Modest' heterogeneity is where each participant has the majority of data from a certain set of classes and other classes make up for the remaining fraction of distribution.
`Modest' heterogeneity is not compared against baseline work, since its behavior is in between severe heterogeneity and homogeneous distribution.
`Extreme' heterogeneity, where the total number of participants is 5, and all the independent classes are equally divided among 5 participants.
`Severe' heterogeneity, where the total number of participants is 10, and all the independent classes are equally divided among 10 participants.
`Homogeneous', where the total number of participants is 10, and all the classes are equally divided among 10 participants.
The values of $r$ and $\delta$ are set to 10 and 0.1, respectively in the experiment for optimal convergence, as decided by empirical observation.

Table \ref{tab:dataset-runs} shows the average accuracy across all the participants under various heterogeneity conditions, on 3 different data sets.
The results shown here are an average of 10 subsequent executions for each case, to avoid the effect of any randomness in data selection for test or train sets.
Table \ref{tab:average-runs} shows the average accuracy across the heterogeneity from Table \ref{tab:dataset-runs}.

\begin{table}[t]
    \centering
    \begin{tabular}{|p{0.8cm}|p{0.8cm}|p{0.8cm}|p{0.8cm}|p{0.8cm}|p{0.8cm}|p{0.8cm}|}
        \hline
        \textbf{Dataset} & \multicolumn{2}{c|}{\textbf{H=Extreme}} & \multicolumn{2}{c|}{\textbf{H=Severe}} & \multicolumn{2}{c|}{\textbf{Homogeneous}} \\
        \hline
        - & pFed-Graph & pFed-Game & pFed-Graph & pFed-Game & pFed-Graph & pFed-Game \\
        \hline
        Fashion MNIST & 0.99 & 0.99 & 0.99 & 0.99 & 0.87 & 0.85 \\
        \hline
        CIFAR-10      & 0.92 & 0.61  & 0.92 & 0.77 & 0.67 & 0.7  \\
        \hline
        CIFAR-100     & 0.54 & 0.57 & 0.56 & 0.55 & 0.31 & 0.01 \\
        \hline
    \end{tabular}
    \caption{Federated model accuracy of pFedGame compared to baseline pFedGraph\cite{ye2023personalized} for heterogeneity of extreme, severe, and homogeneous data distribution among participants.}
    \label{tab:dataset-runs}
\end{table}

\begin{table}[t]
    \centering
    \begin{tabular}{|l|r|r|}
        \hline
        \textbf{Dataset} & \textbf{pFedGraph} & \textbf{pFedGame} \\
        \hline
        Fashion MNIST & 0.95 & 0.94 \\
        \hline
        CIFAR-10  & 0.86 & 0.7 \\
        \hline
        CIFAR-100 & 0.47 & 0.38 \\
        \hline
    \end{tabular}
    \caption{Average of federated model accuracy of pFedGame compared to baseline pFedGraph\cite{ye2023personalized} across heterogeneity of extreme, severe and homogeneous data distribution among participants.}
    \label{tab:average-runs}
\end{table}

From Tables \ref{tab:average-runs} and \ref{tab:dataset-runs}, it is observed that pFedGame performs comparably to state-of-the-art work\cite{ye2023personalized} under \textit{extreme} and \textit{severe} heterogeneity scenarios, for Fashion-MNIST and CIFAR-10 data.
In Fashion-MNIST and CIFAR-10 data, under an extreme heterogeneity scenario of $k=5$, each participant holds 2 classes of the data set. 
Under a severe heterogeneity scenario of $k=10$, each participant holds 1 class of the data set.
But for the CIFAR-100 data set, for $k=5$ and $k=10$, each participant holds 20 classes and 10 classes of the data respectively.
Hence, the local model of each participant in CIFAR-100 data has lower accuracy itself, due to loosely related classes combined in each participant's data distribution.

From table \ref{tab:average-runs} and \ref{tab:dataset-runs}, it is evident that pFedGame performs poorly under homogeneous heterogeneity.
This behavior aligns with the current work's assumption of dynamic graph and personalized data distribution, which has a minor overlap with other participants' data distribution.
With changing data distribution, pFedGame can adapt to variations, since it selects its peers in every federated learning round, and participates in a two-player game to reach convergence, without any presumptions.

\section{Conclusion}

This work introduces `pFedGame' - a novel, game theory-based algorithm for decentralized federated learning, which is suitable for a temporally dynamic network of participants.
The proposed solution is adaptive to many real-world scenarios where having a static central aggregation server is difficult for participants to trust or manage.
`pFedGame' has been compared against the `pFedGraph'\cite{ye2023personalized} method, which is the closest to the current work, and it has shown promising results for heterogeneous data distribution setups.
The solution doesn't perform well for highly homogeneous data distribution among participants, with multiple classes involved in classification tasks. 
In the future, this can further be extended by applying game theory in peer selection for collaboration and adapting the aggregation algorithms for linear and non-linear (deep learning) models.
Decentralized FL has a broad research scope considering the ever-expanding scale of IoT networks and the emergent requirements of personalized edge training from a wide range of intelligent applications.


\bibliographystyle{plain}
\bibliography{refs}

\end{document}